# Donkey and Smuggler Optimization Algorithm: A Collaborative Working Approach to Path Finding

**Ahmed S. Shamsaldin[a]**
ahmed.saadaldin @ukh.edu.krd

**Tarik A. Rashid[a]**
tarik.ahmed@ukh.edu.krd

**Rawan A. Al-Rashid Agha[a]**
rawan.arsn@ukh.edu.krd

**Nawzad K. Al-Salihi[a]**
n.al-salihi@ukh.edu.krd

**Mokhtar Mohammadi[b]**
Mokhtar.mohammadi@uhd.edu.iq

[a]Computer Science and Engineering Department, University of Kurdistan Hewler, Erbil, Kurdistan
[b]Department of Information Technology, University of Human Development, Sulaymaniyah, Iraq.

*Abstract—* Swarm Intelligence is a metaheuristic optimization approach that has become very predominant over the last few decades. These algorithms are inspired by animals' physical behaviors and their evolutionary perceptions. The simplicity of these algorithms allows researchers to simulate different natural phenomena to solve various real-world problems. This paper suggests a novel algorithm called Donkey and Smuggler Optimization Algorithm (DSO). The DSO is inspired by the searching behavior of donkeys. The algorithm imitates transportation behavior such as searching and selecting routes for movement by donkeys in the actual world. Two modes are established for implementing the search behavior and route-selection in this algorithm. These are the Smuggler and Donkeys. In the Smuggler mode, all the possible paths are discovered and the shortest path is then found. In the Donkeys mode, several donkey behaviors are utilized such as Run, Face & Suicide, and Face & Support. Real world data and applications are used to test the algorithm. The experimental results consisted of two parts, firstly, we used the standard benchmark test functions to evaluate the performance of the algorithm in respect to the most popular and the state of the art algorithms. Secondly, the DSO is adapted and implemented on three real-world applications namely; traveling salesman problem, packet routing, and ambulance routing. The experimental results of DSO on these real-world problems are very promising. The results exhibit that the suggested DSO is appropriate to tackle other unfamiliar search spaces and complex problems.

## 1. INTRODUCTION

Swarm Intelligence has been widely used among research communities of diverse backgrounds to solve various optimization tasks. These algorithms are inspired by the behavior of social animals and they are part of the artificial intelligence field. They are a product of designing several agent systems inspired by the cooperative or competitive behaviors of animals or insects' social lives, such as flocks of birds, schools of fish, cats, ants, termites, bees, wasps, etc. These behaviors naturally contribute enormously to the survival of these species. This has intrigued scientific researchers for many years. Individual animals on their own might not be intelligent; nonetheless, within a group, they can collaborate to perform difficult and complicated tasks via simple actions or interaction with the group. Furthermore, most of the characteristics of these social interactions are self-organized, which means that the action can be performed in a decentralized manner. Examples of this include the construction of nests by termites or wasps, and the capability of ants and bees to adapt themselves to their environment (Blum & Li, 2008).

The earliest types of swarm intelligence are very popular and widely used by scientists. They come from many sources including the Genetic Algorithm (GA) (Bonabeau, et al., 1999), Ant Colony Optimization (ACO) (Dorigo et al., 2006), Particle Swarm Optimization (PSO) (Kennedy & Eberhart, 1995), Artificial Bee Colony (ABC) (Karaboga, 2010), Cuckoo search Algorithm (CS) (Yang & Deb, 2009), Bat algorithm (BA) (Yang, 2010), Cat algorithm (Chu et al., 2006) etc. It is also important to mention the key reasons for the popularity of these algorithms and their uses in a wide range of applications. First, these algorithms are so simple to implement as they are mainly reflections of behaviors or representations of some social aspects of a group of animals and their evolutionary processes. Also, they are adaptable to solve different problems and the two most essential elements for a problem to be represented in these algorithms are inputs and outputs. Furthermore, they are mathematically very simple where they do not depend on gradient methods and no mathematical derivations are involved in these types of algorithm. These algorithms also approach problems metaheuristically, this means that the optimizations are performed stochastically—search space derivations are neglected as the optimization in these algorithms initially provide random solutions. Therefore, the optimization is done through iterative processes. Finally, they also avert solutions that are optimal within an adjacent set of candidate solutions as these algorithms hold stochastic characteristics via which the local solutions are avoided and instead the full search space is broadly explored (Mirjalili & Lewis, 2014; Wolpert & Macready, 1997).

The motive behind this paper is that there isn't one global algorithm that can solve competently almost all optimization tasks. This means that a specific algorithm can perform well and produce competent results on some applications, by the same token, the algorithm cannot perform well on other types of applications (Wolpert & Macready, 1997). Thus, this makes it extremely dynamic for improvement, this will help us to introduce our new algorithm called DSO, this algorithm is different from all other previous algorithms in optimization style. The previous algorithms are mainly searching for a global solution, however, if for some reason the global solution has disappeared, and at later stages for some reason, the global solution appears, then the algorithms are not devised or adapted to obtain the best solution, should the condition of the best solutions be found. Therefore, the DSO algorithm has two modes, in the first mode, the algorithm will find the best solution, and in the second mode, the algorithm attempts to maintain the best solutions or to return to the best solution once the conditions are found. Also, the existing algorithms, like ACO have some drawbacks. For instance, the converging time is not certain and the coding is hard and not straightforward (Selvi & Umarani, 2010).

In addition, in all famous nature-inspired global optimization algorithms, such as ant colony, practical swarm, the random technique is used. i.e. they randomly choose a possible solution, test the fitness and set it as the best solution. Then another possible solution is randomly chosen and the fitness is calculated and compared to the fitness of the best solution and so on until the best solution is updated. This is a time-consuming process and the solution that once was the best solution might reappear as the best solution again, but it might take many iterations to get it again as it's a random based process. In DSO, the smuggler evaluates all the possible solutions, in their ideal situation, in one iteration and sequence the solutions based on their fitness then determines the best solution i.e. the best solution chosen is the optimum best solution and none of the other possible solutions can have better fitness than it. In the donkey part, the adaptive part, we try to sustain the best solution and in case it is not good anymore it gets replaced. However, the algorithm keeps the evaluation process running and updates the fitness of the solutions and in case the optimum best solution is back to its fitness, it will be set back to be the best solution. Moreover, the DSO is a population-based algorithm. The population here is the group of solutions that will be used in the adaptive part of the algorithm, hence the donkeys. In other population-based algorithms such as the genetic algorithm, they keep running cross-over and mutation on the different solutions to get the best one. However, in the DSO, there is no need for such processing because the smuggler explores all the solutions and calculates their fitness. Then, these solutions will be put in a population and the best solution will be set from that population. The donkey part will keep evaluating the population and updates the best solution according to the algorithm procedures.

The rest of the paper is organized as follows: Section 2 presents a literature review of swarm intelligence techniques. Section 3 outlines the artificial life. In section 4, the new algorithm is explained in details. The results and discussion of performance evaluation and real applications are presented in Sections 5 and 6, respectively. Finally, Section 7 concludes the work and suggests some directions for future studies.

## 2. LITERATURE REVIEW

With respect to metaheuristics as mentioned earlier, there are two types of metaheuristics in terms of the solutions offered: single solution-oriented and multiple solutions-oriented. Single solution-oriented metaheuristic techniques work on adjusting and enhancing a single candidate solution. Good examples of single solution-oriented of metaheuristics are Simulated Annealing, Iterated Local Search, Variable Neighborhood Search, and Guided Local Search (Kirkpatrick et al., 1983; Blum & Roli, 2003; Talbi, 2009). On the other hand, multiple solutions-oriented metaheuristic techniques attempt to preserve and enhance multiple solutions. Usually, the population features are used for controlling the search. These algorithms start the searching process with a random initial multi-solution and then the population or multi-solution will get improved over many iterations. It is worth mentioning that multiple solutions-oriented metaheuristics are also categorized into Evolutionary Computation and Genetic Algorithms (Mirjalili & Lewis, 2014). Yet, swarm intelligence is another type of population-oriented metaheuristic techniques.

Swarm Intelligence is regarded as a form of decentralized cooperative behavior, relying on self-organized agents in a group (Talbi, 2009). Examples of these are ant colony optimization, particle swarm optimization, social cognitive optimization, penguins search optimization algorithm and artificial bee colony algorithms (Talbi, 2009).

In general, multiple solutions oriented are better than single oriented techniques in having better communication among the individual group. In addition, they tend to work collaboratively to learn about search area that leads them to a better exploration in the search space and not fall into local minima by jumping to better-searching space for a global solution (Mirjalili & Lewis, 2014).

In this paper, we mainly focus and record the previous research works on swarm intelligence as these algorithms are imitating the social behaviors of groups of animals. Dorigo suggested Ant Colony Optimization in 1992, the algorithm mimics the social behavior of ants. Ants are great at determining the shortest path between the nest and food source by using the amount of pheromone that ants use in their search for the shortest path (Dorigo et al., 2006). ACO is used to tackle hard combinatorial optimization tasks. Artificial ants use randomized structure heuristics via which probabilistic decision can be made. The algorithm can demonstrate superior performance when implemented to solve network routing applications, which have unclear (Dorigo et al., 2006; Dorigo and Socha, 2006; Dorigo and Stützle, 2003).

The most common and used algorithm is particle swarm optimization, which was coined by both Kennedy and Eberhart in 1995. This algorithm is inspired by flying birds and fish behavior. The PSO algorithm basically applies many particles that have positions and velocities. The algorithm aims at determining the best particle, which provides the best solution (Kennedy & Eberhart, 1995). PSO is simple for implementation and it has a small number of parameters to modify. It is vigorous and can operate parallel computation as it has high likelihoods to find the global optima and can converge fast. Yet, it has difficulty in defining initial design parameters, thus, it might converge too early and possibly fall into a local minimum, particularly, when solving complex problems (Abdmouleh at el. 2007).

Marriage in Honey Bees Optimization (MBO) was suggested in 2001. This is another swarm intelligence algorithm, which is inspired by the phylogenetic of sociality in Hymenoptera (for examples bees, ants, and wasps) (Abbass, 2001). The algorithm uses the behavior of the mating process in honey-bees. Basically, the key features of very complex types of social organization of some insects are nest construction, cooperation amid adults, covering at least two generation groups, and multiplicative division of labor. The insects that do not have one or two of the aforementioned features are called prosocial and the insects that do not have all the above features are called solitary (Dietz, 1986). MBO has advantages over Genetic Algorithm in performing a local search per iteration. Nonetheless, MBO algorithm would select some random and simple local searching techniques (for example, random and flip walks) via which the chance of getting an optimal solution will be decreased. Thus, the whole performance can be seriously influenced by an agent member that has low competence in the algorithm (Yang et al., 2007).

Artificial fish-swarm algorithm (AFSA), which was developed in 2003 (Li, 2003), is regarded as one of the best optimization approaches within the class of swarm intelligence algorithms. The algorithm is inspired by fish behavior, which is the collective movement of fish. Depending on a succession of natural behaviors, the fish continuously attempt to preserve their gatherings, and therefore, establish intelligent behaviors. Penetrating for food, settlement and avoiding and facing dangers, all occur in a social form, and contacts amongst all fish in a group will produce in intelligent social behavior. AFSA algorithm enjoys several benefits, these are high merging speed, litheness, fault tolerance, and high precision.
,
Monkey Search (Mucherino & Seref, 2007) was proposed in 2007 as a global searching algorithm inspired by the behaviors of monkeys. Monkeys have abilities to climb trees for finding food. The tree twigs are signified as perturbations between two adjacent workable solutions of the given global optimization task. The monkey's climb and descent the trees to spot and update these twigs which lead to better solutions. Monkey Search is able to solve a range of challenging optimization problems, which are containing high dimensionality features, non-differentiability, and nonlinearity with a more rapid convergence rate. The algorithm is particularly easy for implementation because it has a small number of the parameters for modification (Zhao & Tang 2008).

The cuckoo search algorithm is another optimization algorithm coined by Xin-she Yang and Suash Deb in 2009 (Yang & Deb, 2009). The algorithm is inspired by some cuckoo classes via leaving their eggs in other host birds' nests. It was indicated that some host birds might conduct fight with the interfering cuckoos. For instance, when the host bird realizes the eggs are left by other birds, then, the host bird fling these strange eggs away and in case it could not fling them away then it will leave its nest and construct a new nest for its own somewhere else. Previous research work on CS focusing on the optimization problems of discrete or continuous space, yet little work has been conducted on binary problems. Nonetheless, in 2011, (Feng et al., 2014), designed a different of CS combined with a quantum-based method for tackling knapsack problems capably.

Another important algorithm that is commonly used by the researchers is called artificial bee colony; basically, the algorithm has three types of bees; scouts, which are working to explore the search area; and employees and onlookers, which are exploiting the promising solutions. The algorithm mimics the behavior of bees in searching for food sources (Karaboga, 2010). ABC has strong local and global explorations and it has been used for solving several optimization problems. However, it has several parameters that have random initialization and need to be tweaked. In addition, it takes a probabilistic approach in the local search (Yuce et al., 2013).

Bat algorithm is inspired by the bats' behaviors. The most important behaviors of bats are navigation and hunting. They use normal sonar to perform navigation and hunting. The algorithm takes advantages of these behaviors for searching for its prey (Yang, 2010). The algorithm can be easily implemented and is able to search both locally and globally. Also, it can be used for solving many optimization problems. Nonetheless, it has many parameters that can need fine-tuning (Yuce et al., 2013).

In 2012 Krill Herd (KH) algorithm developed and suggested to tackle optimization problems. The KH algorithm depends on the imitation of the herding actions of the population of krill. The least distances of each distinct krill from both the peak concentration of the krill herd and the food substance are measured as the fitness function for the movement of movement (Gandomi, Alavi, 2012). KH might not be able to successfully tackle difficult multimodal functions as it might not succeed in continuing to find better solutions. At this point, Krill Migration operator can spontaneously launch to start again the process (Wang et al., 2014).

Another algorithm was developed in 2014. The algorithm depends on the color shifting behavior of a type of fish called cuttlefish for determining the best solution. This type of fish is famous and it is called cephalopods. It has the facilities to transform its color to either apparently vanish into its environment or to generate spectacular shows. The reflection light from various layers of cells (chromatophores, leucophores, and iridophores) and the amalgamation of some particular cells simultaneously will help cuttlefish to cause a large array of patterns and colors (Eesa et al., 2015).

In addition, a Grey Wolf Optimizer (GWO) as a novel metaheuristic proposed in 2014 for solving optimization problems. This algorithm is inspired by Canis lupus or Grey Wolf. The GWO algorithm imitates the headship and stalking style of these type of wolves in their environment. GWO algorithm uses 4 kinds of grey wolves (alpha, beta, delta, and omega) to represent the headship direction. Furthermore, the algorithm implements hunting in three phases; exploring prey, surrounding prey, and attacking prey (Mirjalili & Lewis, 2014). GWO is simple and not difficult to implement. It has few parameters and does not need derivation information in the initial search. Also, it has a special capability to get the correct stability between the exploration and exploitation in the course of the search, which leads to favorable convergence (Faris et al., 2018).

In 2016, a creative search algorithm named fuzzy harmony search (FHS) was introduced by Peraza et al.,(2016) for solving optimization problems. This recent method uses fuzzy logic for dynamic adaptation of the harmony memory accepting. The purpose of that method is to actively adjust the parameters from the range of 0.7 to 1. There work indicated the effect of using fixed parameters in the harmony search algorithm in addition to using fuzzy logic strategies in order to efficiently tune the parameters (Peraza et al., 2016).

Furthermore, in 2017, the performance of the grey wolf optimizer (GWO) algorithm when a hierarchical operator is introduced in the algorithm was examined (Rodríguez et al., 2017). The new operator is hierarchal that is inspired by the hierarchal social pyramid of the grey wolf. The algorithm is applied to the stimulation of the algorithm in the hunting process and contains 5 different variants. The 5 different variants are as follows: centroid, weighted, etc. The variants were the most effective while using fuzzy logarithm (Rodríguez et al., 2017).

A method using fuzzy logic for dynamic parameter adaptation in the imperialist competitive algorithm was presented in 2017 (Bernal et al., 2017). Firstly, the ICA algorithm was studied in the original form in order to find out how it works and what parameters are more effective regarding the results. Various designs for fuzzy systems for dynamic adjustment of the ICA parameters were proposed as well (Bernal et al., 2017).

In 2018, a new meta-heuristic algorithm was proposed, which is a new bio-inspired optimization algorithm based on the self-defense mechanics of plants (Caraveo, Valdez, and Castillo, 2018). The self-defense mechanics and the techniques are a way for the plants to protect themselves from predators. The algorithm considers the predator-prey model as its basis and it is proposed by Lotka and Volterra. Basically, what this means is when the plant detects the presence of an invading organism, it triggers the emission of chemicals to attract the predator of the invading organism (Caraveo, Valdez & Castillo, 2018).

Another algorithm introduced in 2018 is called a new metaheuristic inspired by the vapour-liquid equilibrium for continuous optimization (Cortés-Toro et al., 2018). In the process of searching for the optimum, the procedure activated the vapor-liquid equilibrium state of multiple binary chemical systems. Each decision variable of the optimization problem behaves as the molar fraction of the lightest component of a binary chemical system. In each system, the equilibrium is altered independently and gradually in two opposing directions and at different rates. Furthermore, for each system, the best thermodynamic conditions of equilibrium are searched and evaluated in order to identify the following step towards the solution of the optimization problem. While the search is being done, incorrect solutions are accepted by the algorithm. This process is done in a controlled way by setting a minimum acceptance probability to restart the exploration in other areas to prevent becoming trapped in locally optimal solutions. In addition, the range of each decision variable is reduced autonomously during the search (Cortés-Toro et al., 2018).

Finally, in 2019, a method for dynamically adjusting parameters in meta-heuristics that are based on integral type 2 fuzzy logic was introduced (Olivas et al., 2019). On the basis of Newton's law of gravity and acceleration, the gravitational search algorithm (GSA) was used to solve optimization problems. However, just like most optimization algorithms, the appropriate adjustment of its parameters is a critical issue. In order to overcome this issue, they used type- 2 fuzzy logic for dynamic parameter adjustment in GSA (Olivas et al., 2019).

All types of swarm intelligence algorithms mentioned above are inspired by different social behaviors of various animals and insects for solving various optimization problems. It is concluded that there isn't any swarm intelligence algorithm in particular that can tackle all optimization problems (Wolpert & Macready, 1997). ACO is very popular amongst the aforementioned algorithms mentioned and it was designed to tackle combinatorial optimization problems. It is used mainly for solving a problem by searching for the shortest path in terms of cost or distance. However, it has some limitations, such as ACO's theoretical analysis is not easy, the distribution of probability alters through iterations, and the time of convergence is not definite (Selvi & Umarani, 2010).

This research work motivates us to formulate a new model that mimics the social behavior of donkeys. The ability of the new algorithm is examined to solve real problems in different areas like packet routing in networking, ambulance routing, traveling salesman problem, road selection in GPS navigation, and any area that involves searching and selecting the best solution among multiple possible solutions. Dealing with critical problems require algorithms that deliver robust, fast, and dynamic solutions because the consequences of not having these might be catastrophic. In DSO, as it will be illustrated in section 6.1 Travel Salesman Problem; who needs to travel around a certain number of cities without repeating the same city, each city will be visited once in each tour, then he goes back to the departure city, we can easily identify how fast the DSO can deliver a verity of solid solutions. To our best knowledge, Swarm intelligence is a collaboration and communication work among usually one species. In our work, we are merging the intelligence, communication, and collaboration among two species (Donkeys and Smuggler) to solve a certain problem. Where in the non-adaptive part of the algorithm, a group of attributes for each possible solution are examined to determine the fitness of them then selecting the one with the best fitness as the best solution. After that, the algorithm will maintain this best solution in its adaptive part using procedures driven from the behaviors of donkeys.

**3. DONKEY'S SOCIAL LIFE**

Donkeys have sets of behaviors that to a certain extent, distinguish them from other animals like having a good memory and ability to learn quickly and easily (The Donkey Sanctuary, 2017). The Donkey Sanctuary (2017) states that the learning ability can be as strong as the one that a dog or a dolphin has. These characteristics made donkeys to be the number one animal for smuggling over the years for both national and international smuggling. Díaz (2015) states that smugglers use horses, mules, and donkeys for conducting their businesses, see Figure 1 below.

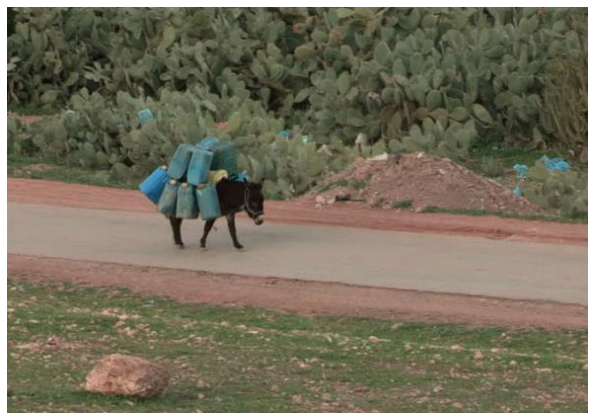

Figure 1: At the Moroccan-Algerian border, donkeys carry gas cans as part of a diesel smuggling (Tapon , 2015)

As Díaz (2015) explains, smugglers ride the horses and load the donkeys as well as the mules with the products that they are smuggling. He also indicates that on familiar paths, donkeys and mules can travel alone without the guidance of the smuggler. Díaz (2015) mentions a story that happened in the Lower Rio Grande Valley where the officers were irritated by a talented donkey (jackass) that could find its way home alone at night. The process was as the following; the smuggler takes the animal across the boarders into Mexico during the day, it gets loaded with liquor during the night and then the donkey is released to go back to its home where the handler is awaiting it (Díaz, 2015).

Another behavior stated by Arab News (2009) is where the donkeys are used to smuggle weed between Yemen and Saudi Arabia. Arab News (2009) reports that the donkeys are trained in homing-pigeon style. Arab News (2009) states that the donkeys are trained in villages in Yemen where the smugglers wear the uniform of the Saudi border police and start hitting the donkeys so they can identify and avoid these uniforms when they see it, i.e. the donkey will change direction or run away once it recognizes the uniform because as, The Donkey Sanctuary (2017) states, donkeys fear and avoid people who are involved in situations that bring pain or fear to the animal.  Also, Arab News (2009) mentions that the donkeys are trained to stop at predetermined locations and wait for someone to unload them and send them back again with the legal load. In the algorithm, this behavior is adopted in the adaptive part, the donkey part, to perform the run action. For instance, if the current network path is dropped or it is not the best path anymore, a run action might be performed to find the new best path and use it.

In addition, donkeys have a high territorial character which enables them to be used, in some countries, for guarding herds of sheep and goats against dogs, foxes, and wolves (The Donkey Sanctuary, 2017; Chan, 2014; Imgur 2014).  The fight/defense techniques of domesticated donkeys are quite simple since they normally live alone or in groups of two (The Donkey Sanctuary, 2017). Therefore, running away is not always the best option for survival whenever a donkey is put in a situation where it senses danger. Its fight/defense behavior is triggered and they use that to save themselves (The Donkey Sanctuary, 2017).

Another behavior that distinguishes donkeys is suicide, donkeys do commit suicide and two cases have been reported in a UN report filed by the Indian Army. Pubby (2008) showed the two cases were donkeys committed suicide after being treated cruelly by their owners. In the first case, an exhausted donkey preferred to be hit to death by his owner instead of dragging a heavily loaded cart through the market. In the second case, a donkey throws itself in the Nile river with its load of water barrel. "A donkey, who had decided to end his miserable and wretched life, ran towards the Nile. As he approached the banks, he plunged into the river and moved towards the current and the strong current of the mighty river swept it to a watery grave," says the report, written by Major Shambhu Saran Singh, posted at the UN mission. "He (the donkey) was still tethered to the water cart he was pulling when he decided to go and drown" (Pubby, 2008). These two behaviors are translated into the algorithm as the face and suicide action. In this action, when the best solution is no longer good, it gets replaced by the second-best solution in the solutions group until it is back to its ideal situation. i.e. if the best path in a network of routers is not the best anymore due to a broken router on the path, then this path will be replaced by the second-best path in the network until the router is fixed.

Furthermore, donkeys have demonstrated the behavior of supporting each other. ABC News (2017) shows, as in Figure 2, a donkey trying to cross a fence but is not able to so he gets help from another donkey who removes a piece of wood to help the herd go through the fence. This behavior is used to create the face and support action in the algorithm. For instance, if the best solution is overloaded, then instead of dropping the solution we can use the second-best solution to support the first one until the load is decreased. In a real-world example, this can be seen as, if a road is congested, then we can use another road to divide the traffic instead of rerouting the whole traffic to another road.

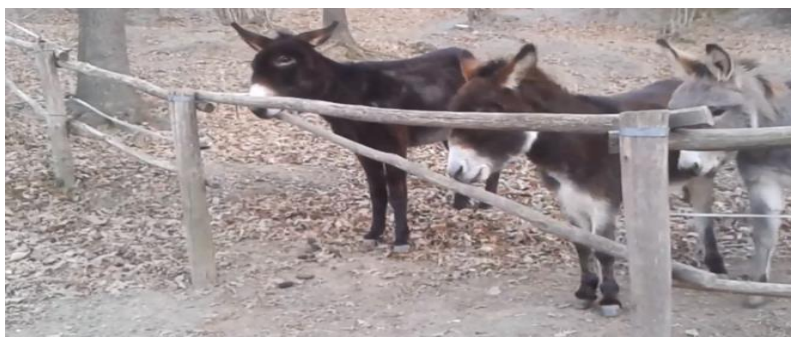

Figure 2: A donkey gets help from another donkey to cross the fence (ABC News, 2017).

In short, the behaviors of the donkey are;
1. Running away from people or events that have caused them a pain in the past.
2. Fight/defense mechanism that is triggered whenever they feel danger.
3. Supporting each other in difficulties or whenever needed.
4. Committing suicide when the situation reaches a level that they can no longer bear to live anymore.

These behaviors along with the smuggler's ones are used to create the DSO algorithm.

# 4. THE ALGORITHM

From the above, the behaviors of donkeys can be concluded in the following points;
1. Run.
2. Face and Support.
3. Face and Suicide.

These behaviors have been translated into our algorithm (See Visuals 1 and 2) where we used the smuggler and the donkey behaviors to construct a two-part algorithm that finds the best solution and react to any changing event in order to maintain that best solution.

*Part I: The smuggler (Non-Adaptive)*

This part is the Non-Adaptive part of the algorithm. This means that this part does not adapt to any changes. In this part, the smuggler will check all the possible routes from the source to the destination then he will decide, based on certain measurements like the time, the safety and the condition of the route, the best route to be taken and the donkey will be sent based on that route. To clarify more, let's consider the following networking example, the parameters, the characteristics, of all the paths will be collected and passed through the smuggler part of the algorithm. In the smuggler part, the fitness calculation is made to find the best route depending on different factors like the cost, the time it takes to reach the destination, i.e. the transmission speed, the bandwidth, delay, and the packet loss. Once this is done, he will send the donkey on the best route. In short, the network operator will be entering the parameters of each path and in the smuggler part, the solutions will be evaluated and the fitness will be calculated. Then, the solutions will be arranged in a group based on their fitness value. The best solution will be set and the donkey will be sent based on that solution.

*Part II: The Donkey (Adaptive)*

On one hand, the Non-Adaptive routing is very good as it is simple and gives good results with relatively consistent topology and traffic. On the other hand, it has a poor performance if the traffic volume or topologies change over time, therefore, this adaptive part of the algorithm is developed.

In the adaptive routing, the decisions are based on the current network state that is measurements/estimates of the topology and the traffic volume. If both the traffic volume and topology or one of these changes, a reaction will be done to avoid losing the path or to avoid a delay from happening. This reaction will be based on the Donkey's behavior.

This is how it will be done, once the user has entered the parameters of each solution, the best solution will be calculated. This will be done in the first part of the algorithm, in the Non-Adaptive part where the routing table will be dealt with as a static table. Then, a choke packet (or any other traffic controlling mechanism) will be sent to update the routing table and this is where the Adaptive part will start. Every time the table is updated by the choke packet, the best solution will be calculated again to update it. This choke packet part will serve as a congestion sensing where we try to fix and avoid losing the path. When the results from the choke packet indicate that there is a drop in the fitness of the best solution or its fitness is not good anymore (another solution has a higher fitness now), one of the following actions will be done;

1. Run: change the path to the other best one (best solution)
   When the best solution that has been determined in the non-adaptive part of the algorithm is no longer the best, it will be dropped and a new best solution will be set according to the new changes.
2. Face and Suicide: fixing the path that we are using (optimizing the best solution). Simultaneously, drop the current path and use the other best solution while fixing the blocked one, there is no recalculation for the fitness of the possible solutions population.
   If the best solution that has been set in the first part of the algorithm is no longer the best due to any changes that affect its fitness and we would like to keep that path. Then we can drop that solution until its back to its ideal conditions and until that happens we can use the second-best solution in the solutions set.

3. Face and Support: when signs of congestion or overloading start to appear in the best solution that was set by the smuggler, we can avoid dropping the solution by assigning the second-best solution in the solutions set to do the same task as the best solution until the signs of the congestion or overloading are gone. i.e. instead of using one channel to transmit and receive data, we can use two channels to reduce congestion or overload. There is no recalculation for the fitness of the possible solutions population.

Flowchart 1, below shows the execution of the algorithm.

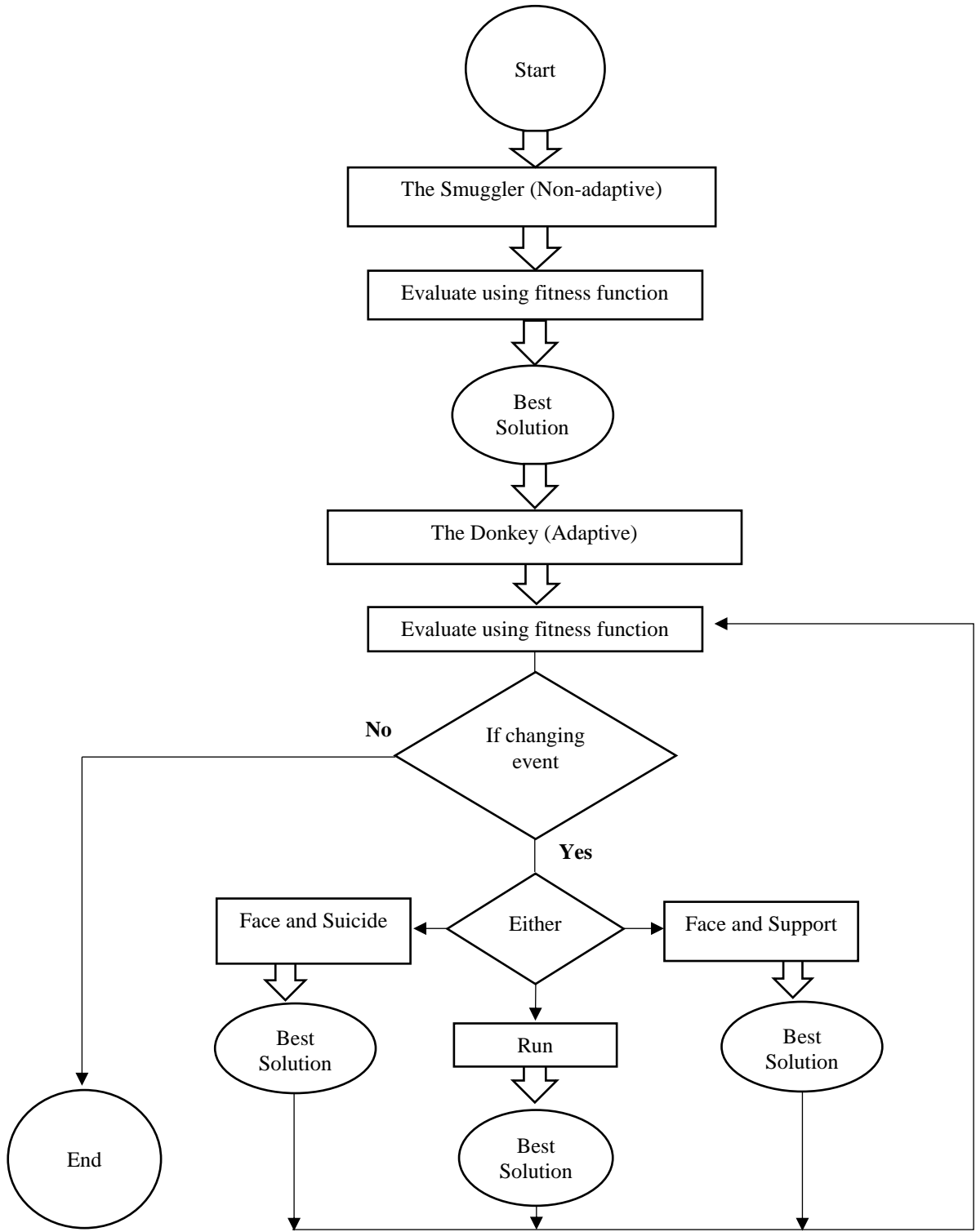

Flowchart 1: The execution of the algorithm

Visual 1: The pseudocode of the DSO Algorithm is described through parts 1 and 2 as shown below.

*Part I: The Smuggler*
*Begin*

    1. Determining the parameters of each solution.

    2. Calculating the fitness of each possible solution using Equation (1):

$$f(x_i) = \frac{\sum_{j=0}^{J} x_{ij} + \prod_{j=0}^{J} x_{ij}}{\sum_{z=0}^{Z} x_{iz} + \prod_{z=0}^{Z} x_{iz}} \quad (1)$$

Where $x_{ij}$ is a possible solution, i is the number of possible solutions, j is the number of parameters of each possible solution that are directly proportional and z is the number of parameters of each possible solution that are reversely proportional. The numerator holds the parameters that are directly proportional and the denominator holds the parameters that are reversely proportional.

    3. Choose the best solution and send the donkey.

*End*

*Part II: The Donkey*
*Begin*

    1. Use the determined solution.
    2. Evaluate the current solution in terms of fitness (keep running the fitness function to find a better solution in case of fitness changing events).
    3. If Yes: (there are signs of congestions):
        *1- Run:* re-evaluate the fitness of the possible solutions population and update the best solution.
        *2. Face and Suicide:* Using the following formula, we can determine the second solution that can be used as a best solution. There is no re-evaluation for the fitness of the possible solutions population.

$$best_{Suicid\ Solution} = f(x_i) - f(best_{Solution}) \quad (2)$$

    Where i is the number of possible solutions.
    The fitness of the best solution is the minimum one in the solutions population therefore, we subtract the fitness of the best solution from the fitness of the possible solutions and the one that gives the minimum difference is considered as the new best solution.

        *3. Face and Support:* When the best solution becomes overloaded, use the second-best solution to support it (the two solutions will be used for the same purpose) until the original best solution is back to normal status. There no re-evaluation for the fitness of the possible solutions population.

$$second\ best_{Solution} = f(best_{Solution}) - f(x_i) \quad (3)$$

$$best_{Support\ Solution} = best_{Solution} + second\ best_{Solution} \quad (4)$$

    Equation (3) will determine which solution to be used as the second-best solution by subtracting the fitness of the possible solutions from the fitness of the best solution. Then, in Equation (4) we join the best solution with the second-best solution to generate the best support solution that uses the two paths to perform one task.

*End*

Visual 2 below shows further implementation details of the DSO.

| **visual 2:** Implementation details of the DSO algorithm |
|---|
| **1: read** input data: [decision variable ranges], n1,n2<br>　　　　n ⟶ dimensions of the solutions matrix (rows by columns)<br>**Smuggler Part**<br>**2: generate** an initial solutions population randomly<br>**3:　for** row=1 to n1<br>**4:　　　for** col=1 to n2<br>**5:　　　　　**parameters(row,col)=randi([decision variable ranges],n1,n2)<br>**6:　　　end**<br>**7:　end**<br>**8:　for** (e=1 to *n1*) **do**<br>**9:　　　　evaluate** the fitness of each solution **(Equation 1)**<br>**10:　　　update** the population of the possible solutions<br>**11: end**<br>**12:　set** the best solution<br>**13: display** the best solution<br>**14: pass** the best solution to the donkey part<br>**Donkey Part**<br>**15: Evaluate** whether a change in the fitness occurred or not<br>**16: if** a change in the fitness of the best solution is less<br>**17:　　Run:** update the best solution **(Equation 2)**<br>**18:　　Face&Suicide:** set the solution with the second-best fitness in the solutions population<br>　　　　　　　　　to be the best solution.<br>**19:　　Face&Suicide:** use the solution with the second-best fitness in the population to<br>　　　　　　　　　support the best solution, i.e. they will be both used for the same<br>　　　　　　　　　purpose. (no updating for the population fitness) **(Equations 3 and 4)**<br>**20: end if** |

## 5. PERFORMANCE EVALUATION

In this section, we will present the benchmark functions that we have used to evaluate the performance of the DSO along with the obtained results.

*5.1 BENCHMARK FUNCTIONS*

We have used 10 benchmark functions to test the performance of the newly proposed algorithm. The used benchmark functions are divided as the following, four unimodal functions, Equations (5)–(8), four multimodal functions, Equations (9)–(12), in these functions, as the number of dimensions increases, the number of local minima increases exponentially and this makes them, for optimization algorithms, a challenging problem (Cortés-Toro et al., 2018). Also, we have used two multimodal functions with fix dimensions Equations (13) and (14). The multimodal functions show the capability to start from local optima and carrying on the search in different parts of the search space and this makes them highly important (Cortés-Toro et al., 2018). The number of dimensions used is D=30. The minimum values of all these functions, as well as the corresponding solutions, are given in Table (1) below. In addition, Figures 3–4 show 3D views of the first set of benchmark functions that have been used in our evaluation (see Figure 5).

TABLE 1.
OPTIMUM VALUES REPORTED FOR THE BENCHMARK FUNCTIONS, WITH THEIR CORRESPONDING SOLUTIONS AND SEARCH SUBSETS (CORTÉS-TORO ET AL., 2018).

| BenFun | SeaSub | Opt | Sol |
|---|---|---|---|
| F1(X) | [−100,100]30 | 0 | [0]30 |
| F2(X) | [−10,10]30 | 0 | [0]30 |
| F3(X) | [−100,100]30 | 0 | [0]30 |
| F4(X) | [−30,30]30 | 0 | [1]30 |
| F5(X) | [−500,500]30 | −12,569.487 | [420.9687]30 |
| F6(X) | [−5.12,5.12]30 | 0 | [0]30 |
| F7(X) | [−32,32]30 | 0 | [0]30 |
| F8(X) | [−600,600]30 | 0 | [0]30 |
| F9(X) | N/A | N/A | N/A |
| F10(X) | N/A | N/A | N/A |
| F11(X) | [−5,5]2 | −1.0316285 | (0.08983, −0.7126) and (−0.08983, 0.7126) |
| F12(X) | N/A | N/A | N/A |
| F13(X) | [−2,2]2 | 3 | (0,−1) |
| F14(X) | N/A | N/A | N/A |
| F15(X) | N/A | N/A | N/A |

Unimodal test functions:

Sphere Function:

$$f_1(x) = \sum_{i=1}^{n} x_i^2 \tag{5}$$

Schwefel's Function No. 2.22:

$$f_2(x) = \sum_{i=1}^{n} |x_i| + \prod_{i=1}^{n} |x_i| \tag{6}$$

Schwefel's Function No. 1.2:

$$f_3(x) = \sum_{i=1}^{n} \left( \sum_{j=1}^{i} x_j \right)^2 \tag{7}$$

Generalized Rosenbrock's Function:

$$f_4(x) = \sum_{i=1}^{n-1} [100(x_{i+1} - x_i^2)^2 + (x_i - 1)^2] \tag{8}$$

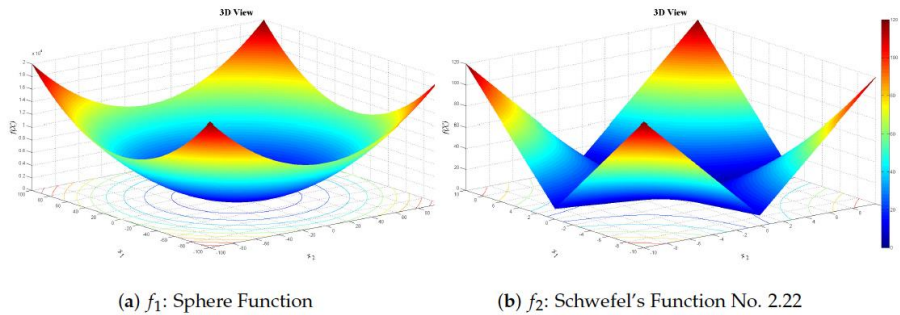

(a) $f_1$: Sphere Function      (b) $f_2$: Schwefel's Function No. 2.22

Figure 3: Cont.

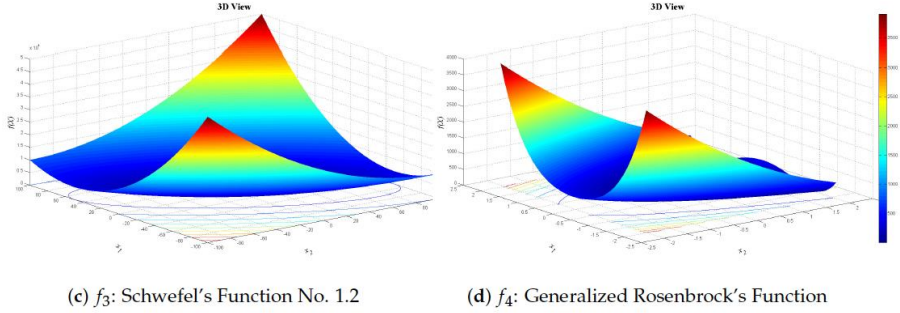

(c) $f_3$: Schwefel's Function No. 1.2    (d) $f_4$: Generalized Rosenbrock's Function

Figure 3: 3D View of the unimodal benchmark mathematical functions (Cortés-Toro et al., 2018).

Multimodal test functions:

Generalized Schwefel's Function No. 2.26:

$$f_5(x) = -\sum_{i=1}^{n}\left(x_i \, sin\left(\sqrt{|x_i|}\right)\right) \tag{9}$$

Generalized Rastrigin's Function:

$$f_6(x) = \sum_{i=1}^{n}[x_i^2 - 10\cos(2\pi x_i) + 10] \tag{10}$$

Ackley's Function:

$$f_7(x) = -20exp\left(-0.2\sqrt{\frac{1}{n}\sum_{i=1}^{n} x_i^2}\right) - exp\left(\frac{1}{n}\sum_{i=1}^{n}\cos(2\pi x_i)\right) + 20 + e \tag{11}$$

Generalized Griewank's Function:

$$f_8(x) = \frac{1}{4000}\sum_{i=1}^{n} x_i^2 - \prod_{i=1}^{n}\cos\left(\frac{x_i}{\sqrt{i}}\right) + 1 \tag{12}$$

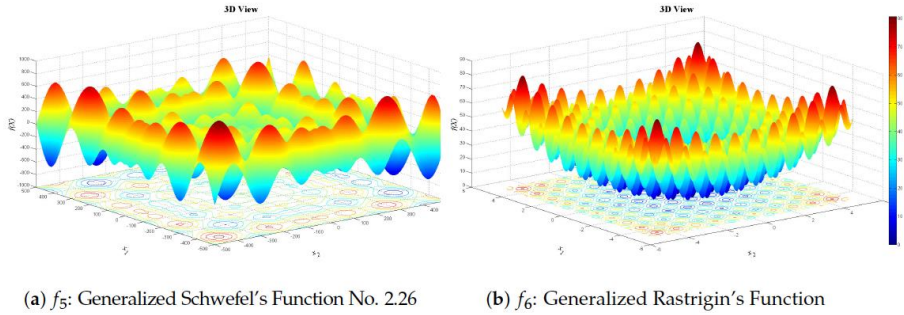

(a) $f_5$: Generalized Schwefel's Function No. 2.26    (b) $f_6$: Generalized Rastrigin's Function

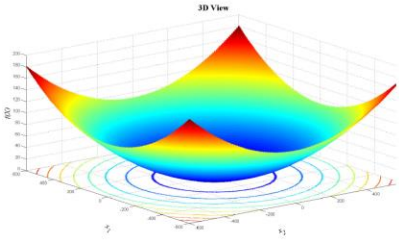

(c) $f_8$: Generalized Griewank's Function

Figure 4: 3D View of some multimodal benchmark mathematical functions (Cortés-Toro et al., 2018).

Multimodal test functions with fixed dimensions:

Six-hump Camel Back Function:

$$f_{11}(x) = 4x_1^2 - 2.1x_1^4 + \frac{1}{3}x_1^6 + x_1x_2 - 4x_2^2 + 4x_2^4 \qquad (13)$$

Goldstein-Price Function:

$$f_{13}(x) = \left[1 + (x_1 + x_2 + 1)^2 \left(19 - 14x_1 + 3x_1^2 - 14x_2 + 6x_1x_2 + 3x_2^2\right)\right] \\ \left[30 + (2x_1 - 3x_2)^2 \left(18 - 32x_1 + 12x_1^2 + 48x_2 - 36x_1x_2 + 27x_2^2\right)\right] \qquad (14)$$

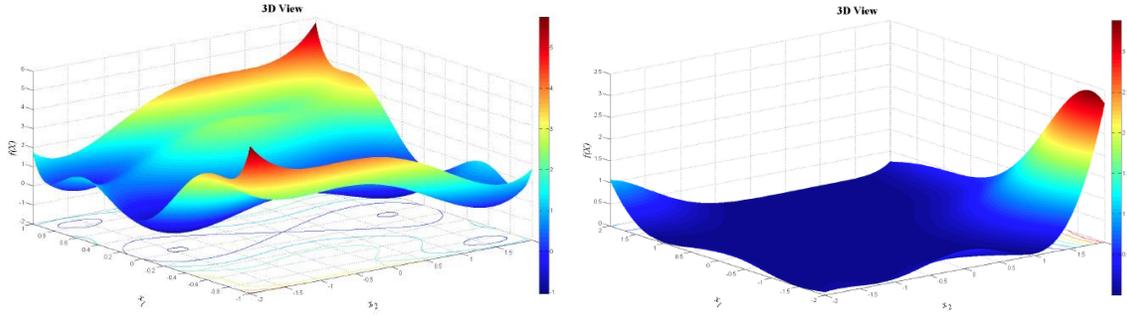

(a) $f_{11}$: Six-hump Camel Back Function    (b) $f_{13}$: Goldstein-Price Function

Figure 5: 3D View of some multimodal benchmark mathematical functions with fix dimensions (Cortés-Toro et al., 2018).

*5.2 RESULTS*

In this section, we will show the results obtained by the DSO and compare it with the published results of VLE, PSO, GSA, DE, and WOA (Cortés-Toro et al., 2018). All the benchmark functions were implemented using MATLAB and DSO parameters were chosen randomly according to the range of each function.
In Table (2) below, we show a comparison between the obtained average fitness (AVG) and the corresponding standard deviation (StdDev) of each benchmark function by the DSO and those that have been obtained by the other algorithms. From the data shown in Table (2) we can notice that the DSO demonstrated very competitive results and compete successfully.

*5.3 COMPARATIVE STUDY*

Table (3) illustrates that, generally, the global average performance of the DSO is 3.3 on a scale of 1 to 6 where being 1 the best and 6 being the worst. This ranking takes into consideration that DSO ranked first twice, second three times, third once, fourth once and sixth three times. In addition, the ranking by the type of the benchmark function is as the following:

- Unimodal functions: 2.5
- Multimodal functions: 2.75
- Multimodal test functions with a fixed number of dimensions: 6

Furthermore, if the global average performance of the DSO is rounded to the nearest integer, as (Cortés-Toro et al., 2018) illustrates, then the DSO ranks third amongst the six algorithms and the evaluated ten benchmark functions.

In short, as (Cortés-Toro et al., 2018) states, there is no algorithm that can perform the best for all optimization problems. Some algorithms will perform very well on some problems, while the others will not perform as well as that algorithm.

TABLE 2.
AVERAGES AND STANDARD DEVIATIONS OBTAINED BY DSO, AND PUBLISHED FOR VLE, PSO, GSA, DE, AND WOA, USING THE CLASSICAL BENCHMARK FUNCTIONS; N=30, (CORTÉS-TORO ET AL., 2018).

| BenFun | Statistic | DSO | VLE | PSO | GSA | DE | WOA |
|---|---|---|---|---|---|---|---|
| $F_1$ | Avg | $6.1402 \times 10^{-10}$ | $4.4989 \times 10^{-7}$ | $1.3600 \times 10^{-4}$ | $2.5300 \times 10^{-16}$ | $8.2000 \times 10^{-14}$ | $1.4100 \times 10^{-30}$ |
|  | StdDev | $3.9 \times 10^{-3}$ | $1.413 \times 10^{-6}$ | $2.0200 \times 10^{-4}$ | $9.6700 \times 10^{-17}$ | $5.9000 \times 10^{-14}$ | $4.9100 \times 10^{-30}$ |
| $F_2$ | Avg | 0.0000 | $3.0840 \times 10^{-6}$ | $4.2144 \times 10^{-2}$ | $5.5655 \times 10^{-2}$ | $1.5000 \times 10^{-9}$ | $1.0600 \times 10^{-21}$ |
|  | StdDev | 0.0000 | $6.0498 \times 10^{-6}$ | $4.5421 \times 10^{-2}$ | 0.19407 | $9.9000 \times 10^{-10}$ | $2.3900 \times 10^{-21}$ |
| $F_3$ | Avg | $8.6688 \times 10^{-4}$ | 5.2020 | 70.126 | $8.9653 \times 10^{2}$ | $6.8000 \times 10^{-11}$ | $5.3900 \times 10^{-7}$ |
|  | StdDev | $1.7 \times 10^{-3}$ | 0.79863 | 22.119 | $3.1896 \times 10^{2}$ | $7.4000 \times 10^{-11}$ | $2.9300 \times 10^{-6}$ |
| $F_4$ | Avg | 0.961 | 79.199 | 96.718 | 67.543 | 0.0000 | 27.866 |
|  | StdDev | $2.8 \times 10^{-3}$ | 37.400 | 60.116 | 62.225 | 0.0000 | 0.76363 |
| $F_5$ | Avg | $-1.3448 \times 10^{-4}$ | $-1.2566 \times 10^{4}$ | $-4.8413 \times 10^{3}$ | $-2.8211 \times 10^{3}$ | $-1.1080 \times 10^{4}$ | $-5.0808 \times 10^{3}$ |
|  | StdDev | $1.7 \times 10^{-3}$ | 68.705 | $1.1528 \times 10^{3}$ | $4.9304 \times 10^{2}$ | $5.7470 \times 10^{2}$ | $6.9580 \times 10^{2}$ |
| $F_6$ | Avg | 0.0000 | 34.5830 | 46.704 | 25.968 | 69.200 | 0.0000 |
|  | StdDev | 0.0000 | 17.8860 | 11.629 | 7.4701 | 38.800 | 0.0000 |
| $F_7$ | Avg | $3.6311 \times 10^{-4}$ | 3.1704 | 0.27602 | $6.2087 \times 10^{-2}$ | $9.7000 \times 10^{-8}$ | 7.4043 |
|  | StdDev | $1.6 \times 10^{-3}$ | 3.9211 | 0.50901 | 0.23628 | $4.2000 \times 10^{-8}$ | 9.8976 |
| $F_8$ | Avg | $4.8168 \times 10^{-7}$ | 0.50737 | $9.2150 \times 10^{-3}$ | 27.702 | 0.0000 | $2.8900 \times 10^{-4}$ |
|  | StdDev | $1.9 \times 10^{-3}$ | 0.50405 | $7.7240 \times 10^{-3}$ | 5.0403 | 0.0000 | $1.5860 \times 10^{-3}$ |
| $F_9$ | Avg | N/A | 0.23693 | $6.9170 \times 10^{-3}$ | 1.7996 | $7.9000 \times 10^{-15}$ | 0.33968 |
|  | StdDev | N/A | 0.28773 | $2.6301 \times 10^{-2}$ | 0.95114 | $8.0000 \times 10^{-15}$ | 0.214864 |
| $F_{10}$ | Avg | N/A | 0.99800 | 3.6272 | 5.8598 | 0.99800 | 2.1120 |
|  | StdDev | N/A | $2.5294 \times 10^{-7}$ | 2.5608 | 3.8313 | $3.3000 \times 10^{-16}$ | 2.4986 |
| $F_{11}$ | Avg | $-1.5718 \times 10^{-6}$ | $-1.0315$ | $-1.0316$ | $-1.0316$ | $-1.0316$ | $-1.0316$ |
|  | StdDev | $3.2 \times 10^{-3}$ | $1.8408 \times 10^{-4}$ | $6.2500 \times 10^{-16}$ | $4.8800 \times 10^{-16}$ | $3.1000 \times 10^{-13}$ | $4.2000 \times 10^{-7}$ |
| $F_{12}$ | Avg | N/A | 0.39815 | 0.39789 | 0.39789 | 0.39789 | 0.39791 |
|  | StdDev | N/A | $4.5697 \times 10^{-4}$ | 0.0000 | 0.0000 | $9.9000 \times 10^{-9}$ | $2.7000 \times 10^{-5}$ |

| F$_{13}$ | Avg | 20.1881 | 3.0097 | 3.0000 | 3.0000 | 3.0000 | 3.0000 |
| | StdDev | 4.2 x 10$^{-3}$ | 1.6256 × 10$^{-2}$ | 1.3300 × 10$^{-15}$ | 4.1700 × 10$^{-15}$ | 2.0000 × 10$^{-15}$ | 4.2200 × 10$^{-15}$ |
| F$_{14}$ | Avg | N/A | −3.8628 | −3.8628 | −3.8628 | N/A | −3.8562 |
| | StdDev | N/A | 6.6880 × 10$^{-5}$ | 2.5800 × 10$^{-15}$ | 2.2900 × 10$^{-15}$ | N/A | 2.7060 × 10$^{-3}$ |
| F$_{15}$ | Avg | N/A | −3.3179 | −3.2663 | −3.3178 | N/A | −2.9811 |
| | StdDev | N/A | 2.1311 × 10$^{-2}$ | 6.0516 × 10$^{-2}$ | 2.3081 × 10$^{-2}$ | N/A | 0.37665 |

TABLE 3.
RANKING OF THE OPTIMIZATION RESULTS (AVERAGE FITNESS) OBTAINED APPLYING DSO, VLE, PSO, GSA, DE, AND WOA TO THE CLASSICAL BENCHMARK FUNCTIONS CONSIDERED; N = 30, (CORTÉS-TORO ET AL., 2018).

| BenFun | 1st | 2nd | 3rd | 4$^{th}$ | 5th | 6$^{th}$ | Rank | Subtotal |
|---|---|---|---|---|---|---|---|---|
| F$_1$ | WOA | GSA | DE | DSO | VLE | PSO | 4 | |
| F$_2$ | DSO | WOA | DE | VLE | PSO | GSA | 1 | |
| F$_3$ | DE | WOA | DSO | VLE | PSO | GSA | 3 | |
| F$_4$ | DE | DSO | WOA | GSA | VLE | PSO | 2 | 10 |
| F$_5$ | VLE | DE | WOA | PSO | GSA | DSO | 6 | |
| F$_6$ | DSO, WOA | GSA | VLE | PSO | DE | | 1 | |
| F$_7$ | DE | DSO | GSA | PSO | VLE | WOA | 2 | |
| F$_8$ | DE | DSO | WOA | PSO | VLE | GSA | 2 | 11 |
| F$_{11}$ | PSO, GSA, DE, WOA | VLE | | | | DSO | 6 | |
| F$_{13}$ | PSO, GSA, DE, WOA | VLE | | | | DSO | 6 | 12 |
| | | | | | | **Total:** | 33 | |
| | | | | | | **overall rank:** | 33/10=3.3 | |
| | | | | | | **F1- F4:** | 10/4=2.5 | |
| | | | | | | **F5- F8:** | 11/4=2.75 | |
| | | | | | | **F11&F13:** | 12/2=6 | |

## 6. APPLICATIONS

Generally, the algorithm can be used in any area that involves determining the best solution out of multiple possible solutions as mentioned before. The problems where we have simulated the algorithm are travel salesman problem, packet routing, and ambulance routing.

*6.1 TRAVELING SALESMAN PROBLEM*

One of the famous benchmarks for new techniques in combinatorial optimization is the traveling salesman problem (TSP) (Yang et al., 2008-2007). Simply, the problem is a salesman who needs to travel around a certain number of cities with no repetition, each city will be visited once in each tour, then he goes back to the departure city (Yang et al., 2008-2007). The question of the travel salesman problems is, in which order should the cities be visited so that the distance traveled is minimized (Yang et al., 2008-2007).

The DSO algorithm has been applied on the TSP, using MATLAB see Appendix 1, like the following; for one iteration, the smuggler in the first part of the algorithm examines all the distances between the cites and according to his experience he will determine the shortest path to be taken to each city. This is repeated for every iteration to determine the shortest path to be taken to visit all the cities starting from different cities. In the second part of the algorithm, after pointing out the departing city and the shortest path to follow to visit all the cities, new possible paths will be searched and suggested, in case the shortest path determined in the first part is not available anymore for any reason, to be taken according to the reaction chosen in the DSO algorithm.

In (Santosa, 2012), it is shown how the ACO was applied to the TSP problem. ACO was applied as the following:
- Place the ants in the cities randomly.
- For each ant:
    1) Pick a city that has not been visited yet until the visit is finished.
    2) Optimize the visit.
    3) Update pheromone using the ACO formula using $\tau_{ij} = \tau_{ij} + 1$ *length (tour)*     (15)

- Evaporate Pheromone   $\tau_{ij=}(1-\rho)*\tau_{ij}$     (16)

Santosa (2012) studied an example for a TSP with five cities. The distances between the cities were given by the matrix *d* shown below and the number of the ants is three.

$$d = \begin{matrix} 0 & 10 & 12 & 11 & 14 \\ 10 & 0 & 13 & 15 & 8 \\ 12 & 13 & 0 & 9 & 14 \\ 11 & 15 & 9 & 0 & 16 \\ 15 & 8 & 14 & 16 & 0 \end{matrix}$$

The solution as (Santosa, 2012) solved is as stated below.
**Step one:** find the visibility *(h)* by taking the inverse of the distance *1/d*. Then, they gave the same value of pheromone, one, to all cities.
**Step two:** calculating the probability of which city to move to consider city one as the departure city.
**Step three:** a random number is generated to decide the order of the cities is visited by comparing it with the cumulative value of the probability.
**Step four:** the addition and evaporation of the pheromones are calculated.
Step two to four is repeated until reaching the maximum number of iterations.
Santosa (2012) showed that in the first iteration, where city one is the departure city, they found three routes with three weights:

*Route 1: 1 - 4 - 3 - 5 - 2 - 1    weight: 52*
*Route 2: 1 - 4 - 2 - 5 - 3 - 1    weight: 60*
*Route 3: 1 - 4 - 5 - 2 - 3 -1     weight: 60*

Then after reaching the maximum number of iterations, they found out that route 1 is the best route to be taken by the traveling salesman.

When DSO was applied on the same TSP example as mentioned earlier, the first part of the algorithm considered each city as a departing city once and all the iterations were performed to find out all the possible routes and their weights and the shortest path to be followed was calculated lastly, see Figure 6 below. The whole process of part one took around 0.0285 seconds:

*Path from city 1: 1   2   5   3   4   1    weight = 52*
*Path from city 2: 2   5   3   4   1   2    weight = 52*
*Path from city 3: 3   4   1   2   5   3    weight = 52*
*Path from city 4: 4   3   1   2   5   4    weight = 55*
*Path from city 5: 5   2   1   4   3   5    weight = 52*

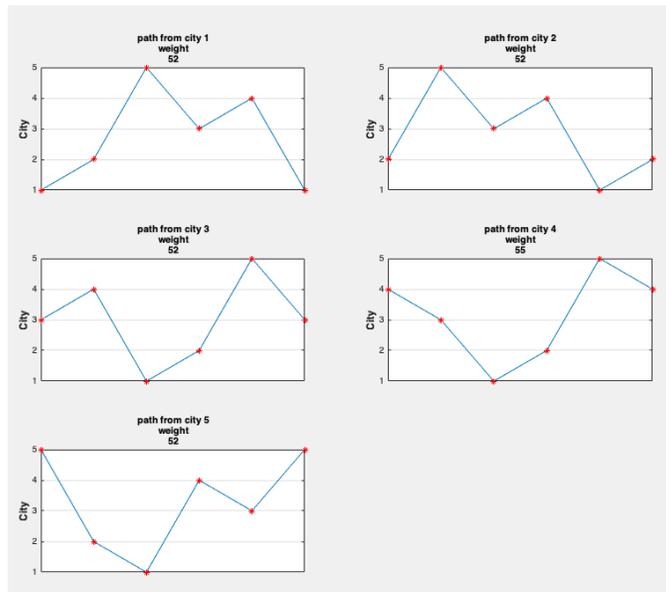

Figure 6: Shortest paths to be taken from every city to visit all the other cities and finish the salesman tour.

Part two of the algorithm, which aims at maintaining the route to all cities in cases where the shortest path chosen in the first part is no longer available. In this part, after determining the departing city and in addition to the shortest route, the algorithm scans the whole dataset and calculates all the other possible routes from that city to visit all the cities in the set as well as their weights, see Figure 7. The execution time of this part was around 0.010 seconds.

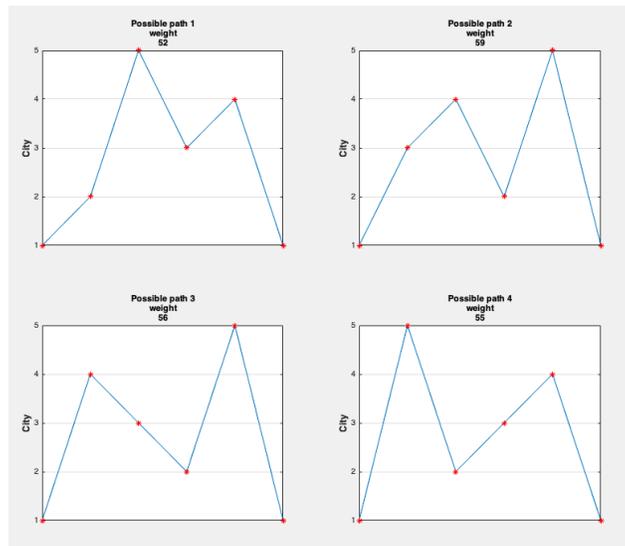

Figure 7: The shortest path as well as the other possible paths that can be taken from city one to visit all the other cities.

When the ACO was applied with three ants, it gave one best solution and the weight of each solution changed with each time the code was executed, it ranged from 52 to 62. The results of the multiple runs, 16 executions with 100 iterations for each execution, with execution time being between 0.037442 to 0.041263, were as the following;

```
Run 1: bestrute = 1   2   3   4   5   1   mincost = 62
Run 2: bestrute = 1   2   5   4   3   1   mincost = 55
Run 3: bestrute = 1   2   3   4   5   1   mincost = 62
Run 4: bestrute = 1   2   3   4   5   1   mincost = 62
Run 5: bestrute = 1   2   5   3   4   1   mincost = 52
Run 6: bestrute = 1   4   3   5   2   1   mincost = 52
Run 7: bestrute = 1   4   3   5   2   1   mincost = 52
Run 8: bestrute = 1   2   5   4   3   1   mincost = 55
Run 9: bestrute = 1   2   4   3   5   1   mincost = 62
Run 10: bestrute = 1   4   3   2   5   1 mincost = 55
```

```
Run 11: bestrute = 1   4   2   5   3   1 mincost = 60
Run 12: bestrute = 1   4   3   2   5   1 mincost = 55
Run 13: bestrute = 1   2   5   4   3   1 mincost = 55
Run 14: bestrute = 1   3   4   5   2   1 mincost = 55
Run 15: bestrute = 1   2   5   4   3   1 mincost = 55
Run 16: bestrute = 1   3   2   5   4   1 mincost = 60
```

The best solution for this distance matrix is *1   2   5   3   4   1* with the weight of *52*, from the execution results above, we can notice that it appears 3 times out of 16 executions and that wasn't stable, it might appear from the first execution and it might not. In addition, we can see that in all the solutions given by the ACO, the starting city is always the first city i.e. it doesn't consider the other cities as a starting point. In order to make the results more stable, we increased the number of ants to 10 and the results came as the following, with execution time between 0.11 to 0.13;

```
Run 1:  bestrute = 1   4   3   2   5   1  mincost = 55
Run 2:  bestrute = 1   4   3   5   2   1  mincost = 52
Run 3:  bestrute = 1   2   5   3   4   1  mincost = 52
Run 4:  bestrute = 1   3   4   2   5   1  mincost = 58
Run 5:  bestrute = 1   2   5   3   4   1  mincost = 52
Run 6:  bestrute = 1   2   5   4   3   1  mincost = 55
Run 7:  bestrute = 1   4   3   2   5   1  mincost = 55
Run 8:  bestrute = 1   4   3   2   5   1  mincost = 55
Run 9:  bestrute = 1   2   5   3   4   1  mincost = 52
Run 10: bestrute = 1   2   5   3   4   1  mincost = 52
Run 11: bestrute = 1   2   5   3   4   1  mincost = 52
Run 12: bestrute = 1   2   5   3   4   1  mincost = 52
Run 13: bestrute = 1   2   5   3   4   1  mincost = 52
Run 14: bestrute = 1   4   3   5   2   1  mincost = 52
Run 15: bestrute = 1   4   3   5   2   1  mincost = 52
Run 16: bestrute = 1   4   3   2   5   1  mincost = 55
```

From the results above we can see that the best solution is appearing 10 times out of 16 executions. So, we can conclude that increasing the number of ants will give better results and will increase the execution time.

When the DSO was applied to solve the TSP, in one run, the first part of the algorithm gave all the possible paths and their weights, with an execution time of 0.0285 seconds;

*Path from city 1: 1   2   5   3   4   1     weight = 52*
*Path from city 2: 2   5   3   4   1   2     weight = 52*
*Path from city 3: 3   4   1   2   5   3     weight = 52*
*Path from city 4: 4   3   1   2   5   4     weight = 55*
*Path from city 5: 5   2   1   4   3   5     weight = 52*

We can see that the DSO calculated the best path to start with from each city and one can notice that there are four options to start with;

*Path from city 1: 1   2   5   3   4   1     weight = 52*
*Path from city 2: 2   5   3   4   1   2     weight = 52*
*Path from city 3: 3   4   1   2   5   3     weight = 52*
*Path from city 5: 5   2   1   4   3   5     weight = 52*

Once a path is chosen, the DSO will calculate, with an execution time of 0.010 seconds, all the other possible paths that can be used to replace the chosen path in case that chosen path is no longer fitted.

In case path from city one is chosen, the other possible paths will be:

```
Path 1 = 1   2   5   3   4   1 weight = 52
Path 2 = 1   3   4   2   5   1 weight = 59
Path 3 = 1   4   3   2   5   1 weight = 56
Path 4 = 1   5   2   3   4   1 weight = 55
```

In case path from city two is chosen, the other possible paths will be:

```
Path 1 = 2   1   4   3   5   2 weight = 52
Path 2 = 2   3   4   1   5   2 weight = 55
```

```
                        Path 3 = 2   4   3   1   5   2 weight = 58
                        Path 4 = 2   5   3   4   1   2 weight = 52
```

In case path from city three is chosen, the other possible paths will be:

```
                        Path 1 = 3   1   2   5   4   3 weight = 55
                        Path 2 = 3   2   5   1   4   3 weight = 56
                        Path 3 = 3   4   1   2   5   3 weight = 52
                        Path 4 = 3   5   2   1   4   3 weight = 52
```

In case path from city five is chosen, the other possible paths will be:

```
                        Path 1 = 5   1   2   3   4   5 weight = 63
                        Path 2 = 5   2   1   4   3   5 weight = 52
                        Path 3 = 5   3   4   1   2   5 weight = 52
                        Path 4 = 5   4   3   1   2   5 weight = 55
```

From the above, we can conclude that the DSO gives more and stable options, in around 0.0385 seconds than the ACO which gives unstable options for the same distance matrix with an execution time of 0.11 to 0.13 seconds for each execution.

## 6.2 PACKET ROUTING

Designing any network for validating this algorithm, in particular, may not give the best results. As this algorithm takes five parameters in consideration and the parameters for any network would be the cost, bandwidth, delay, transmission speed, and packet loss. However, choosing an algorithm that finds the shortest path won't take all these parameters together into consideration. Therefore, a network has been designed and configured using the OSPF (Open shortest path first) routing protocol. As Deng et al.,(2014) defines OSPF is an interior gateway protocol according to RFC 2328, which is used to distribute routing information within an Autonomous System. The reason for choosing OSPF is that OSPF works by link-state technology and use SPF algorithms, which were developed by Dijkstra's for calculating the shortest path and to prove that the DOS algorithm works fine when it is compared. The link state technology is able to build the entire topology for a network and then calculate the best path from all links in the network. In other words, it has a complete overview of the network (Cisco, 2005). Other advantages in choosing OSPF are that there is no limitation on hop count unlike RIP as well as having better convergence than RIP. This is due to routing changes are propagated instantly rather than periodically (Cisco, 2005).

A brief comparison between the routing protocols can be seen in Table 4 including the OSPF routing protocol.

TABLE 4
COMPARISON BETWEEN ROUTING PROTOCOLS

|  | **RIO v1** | **RIP v2** | **IGRP** | **EIGRP** | **OSPF** | **IS-IS** | **BGP** |
|---|---|---|---|---|---|---|---|
| Interior/Exterior | Interior | Interior | Interior | Interior | Interior | Interior | Exterior |
| Type | Distance Vector | Distance Vector | Distance Vector | Hybrid | Link-state | Link-state | Path Vector |
| Default Metric | Hopcount | Hopcount | Bandwidth/Delay | Bandwidth/Delay | Cost | Cost | Multiple Attributes |
| Administrative Distance | 120 | 120 | 120 | 90 internal 170 external | 110 | 115 | 20 internal 200 external |
| Hopcount Limit | 15 | 15 | 225 | 224 | None | None | - |
| Coverage | Slow | Slow | Slow | Very fast | fast | fast | Average |
| Update timers | 30 sec | 30 sec | 90 sec | Only when changes occur | Only when changes occur, Table is refreshed every 30 minutes | Only when changes occur | Only when changes occur |
| Updates | Full table | Full table | Full table | Only changes | Only changes | Only changes | Only changes |
| Classless | No | Yes | No | Yes | Yes | Yes | Yes |
| Support VLSM | No | Yes | No | Yes | Yes | Yes | Yes |
| Algorithm | Bellman-Ford | Bellman-Ford | Bellman-Ford | Dual | Dijkstra | Dijkstra | Best Path Algorithm |

Furthermore, the Dijkstra Algorithm was developed by the Dutch computer scientist Edsger Dijkstra. As mentioned earlier the shortest path calculated by using the Dijkstra algorithm, which is a quite complicated algorithm. Therefore, the Dijkstra algorithm simply is that it assumes each router in the network as head of the tree and calculates the shortest path for each router based on the accumulative cost required to reach the routers. All routers will use the same link-state database hence all routers have an overview of the network topology (Venkat, 2014).

To validate that the DSO algorithm is, in fact, working in finding the best solution, a network topology was designed using cisco packet tracer. The topology is a partial mesh network as the devices are connected with many redundant interconnections between the nodes. Additionally, in a partial mesh network if one node is no longer operational, the rest of the nodes are still able to communicate; either directly or through an intermediate or multiple ones (Sparrow, 2017). This network is created to match Dijkstra`s Algorithm, so, 8 routers were configured and connected with each other to create three different paths as well as using two PCs to ping through the network. Furthermore, the OSPF routing protocol was applied to create the routing tables as OSPF can choose the best path. After the router configurations, the OSPF was enabled on the routers using "router OSPF <process-id>" command; the process ID is simply a numeric value local to the router. Then the next step is defining the IP address on which the OSPF runs and assigning the area ID to that interface was done by using the "network <IP address> <mask> <area-id>" command. The mask is needed because it is used as a shortcut since it helps in setting a list of interfaces in the same area with one configuration line. In addition, the mask contains the wildcard mask where "0" means a match and "1" is a "do not care" bit, in this network 0.0.0.3 was the wildcard mask. In this network the connection is a point-to-point link as a router is connected to another router, therefore, the subnet /30 is used and for this network; the Netmask is 255.255.255.252 hence the Wildcard mask is 0.0.0.3. As mentioned before the OSPF is depending on the metric cost and by default the cost of an interface in Cisco packet tracer is calculated based on the entered bandwidth. However, the cost of an interface can be forced with the command "is OSPF cost interface cost_value" which was used to determine different costs for the three paths. As seen in Figure 8 the three paths have different costs and a number of nodes and the path with the least cost is the best path according to OSPF. Thus, the 3rd path has the least cost with 2000 from router 0 passing through routers 3-4-7 to router 5. Figure 9 is the tracing route for the shortest path for the first designed network;

Through such a network the needed parameters (Cost, Delay, Packet loss, Bandwidth and Transmission Speed) can be jotted down through pings. Moreover, Figure 10 shows the ping for the shortest path as 4 packets are being sent with no loss with approximate round-trip time in milliseconds.

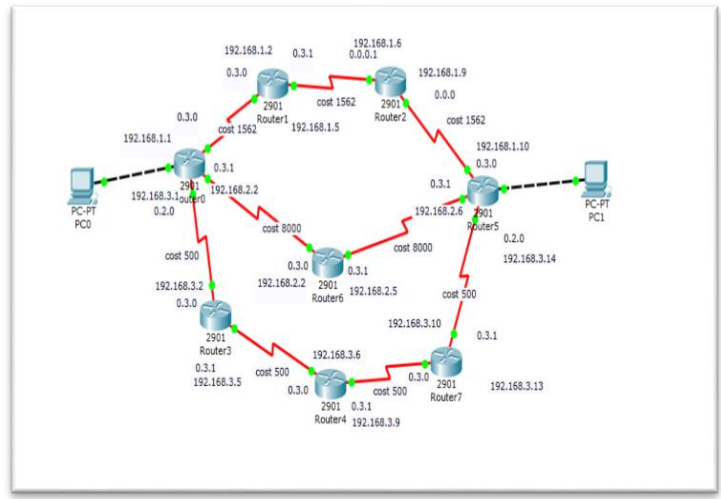

Figure 8: First designed network using OSPF Protocol.

```
Command Prompt

Tracing route to 20.0.0.2 over a maximum of 30 hops:

  1    1 ms      0 ms      0 ms      10.0.0.1
  2   12 ms      1 ms      0 ms      192.168.3.2
  3   11 ms     12 ms     12 ms      192.168.3.6
  4   14 ms     12 ms     25 ms      192.168.3.10
  5   32 ms     29 ms     18 ms      192.168.3.14
  6   28 ms     18 ms     27 ms      20.0.0.2

Trace complete.
```

Figure 9: Tracing route for the shortest path.

```
Command Prompt
Pinging 192.168.1.10 with 32 bytes of data:

Reply from 192.168.1.10: bytes=32 time=62ms TTL=251
Reply from 192.168.1.10: bytes=32 time=16ms TTL=251
Reply from 192.168.1.10: bytes=32 time=17ms TTL=251
Reply from 192.168.1.10: bytes=32 time=13ms TTL=251

Ping statistics for 192.168.1.10:
    Packets: Sent = 4, Received = 4, Lost = 0 (0% loss),
Approximate round trip times in milli-seconds:
    Minimum = 13ms, Maximum = 62ms, Average = 26ms
```

Figure 10: Ping of the shortest path.

The data were taken for each path and can be seen in Table 5 below:

TABLE 5
DATA OF EACH PARAMETER FOR THE THREE PATHS

| Solution/Parameter | P. Loss | P. Delay | Cost | Bandwidth | Transmission Speed | Fitness $f(x)$ |
|---|---|---|---|---|---|---|
| $X_1$ | 0 | 70 | 5186 | 1544 | 15 | 0.0894931637380864 |
| $X_2$ | 0 | 55 | 26062 | 1544 | 12 | 0.00353724795155033 |
| $X_3$ | 0 | 19 | 4062 | 1544 | 16 | 0.14599188313114900 |

The data for each parameter was needed for the first two paths as the network can only give for the shortest path with pinging. That's why; pings were made between each node to force the packet to go through each of the first and second paths. The results are tabulated and can be seen above for all the three paths with third being the shortest path as it has the least cost. When these data were run through DSO, the following can be concluded, since the packet loss is zero for all the paths and the bandwidth is also the same for all paths, these two parameters were ignored by the algorithm.

The packet delay is the lowest on the third path, then, the second path, then, the first path, therefore; the third path is the best in this one. The cost is the lowest on the third path, then, the first path, then, the second path, thus, the third path is again the best one. Finally, when we consider the transmission speed, the second path is the best, then, the first one then the third one, however, the DSO still chose the third path, X3, as the best solution and that is due to the fact that the difference in the speed is not as big as the difference in the cost. If the second path is chosen as the best solution, then it wouldn't be very sufficient because the cost will be dramatically high for a slightly faster path. The user will be paying 22000 IQD more for an only four milliseconds faster path and this is not really practical.

If the user wants the fastest path without considering the cost then it can be easily done by giving a zero to the cost parameter for each path and the algorithm with determining the best solution based on the other two parameters, the packet delay, and transmission speed. This shows that the algorithm is flexible, it doesn't require five parameters to operate, and it can operate and give the best solution even if only one parameter is given but the results will be remarkably more accurate when the five parameters are given.

The DSO will maintain the third path as the first solution but in case of any changes that affect the fitness of that solution, one of the behaviors of the adaptive part of the algorithm will be implemented, the second-best solution to be chosen is the first path. Even though the second solution is a bit faster and the delay is less but again the cost of that path is extremely high and the speed and delay differences are not high enough to worth this great cost.

The 2nd design, as shown in Figure 11, has different paths and costs and the parameters of this design have been tested on the DSO algorithm and gave the same result. Furthermore, in a partially connected network, it is possible that some nodes are connected to exactly one other node; yet some other nodes can be connected to two or more other nodes by means of a point-to-point link. This exploits redundancy of the mesh topology that is physically connected, without having the expenditure and complication needed to connect between each node in the network (Rouse, 2014). As such a connection can be seen within the second design and it is as well configured with OSPF routing protocol but with different costs for each link. Again, the cost was enforced by the command "is OSPF cost interface cost value". The new routing table chooses the path according to the least cost on the path while there is no less administrative distance than the default value for OSPF. The best path is the first one between routers (0, 6, 7 and 4) as shown in Figure 11.

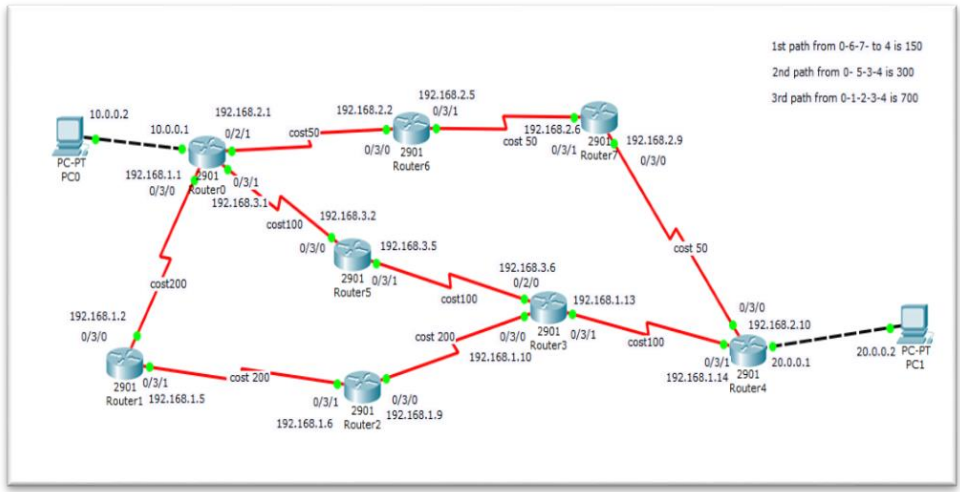

Figure 11: Second designed network topology.

Figure 12 below shows the tracing route indicating the shortest path also with the ping for the shortest path;

Figure 12: Tracing route and the ping for the shortest path

The data for each parameter was needed for all three paths and the network gives only the shortest path when it is pinged. Therefore, pings were made between each node as well to force the packet to go through each of the second and third paths. The results are tabulated below and can be seen the first path is the shortest path as it has the least cost with data for the other two paths as well.

From the data in Table 6, we can make the same remarks that have been made on the previous data set. The packet loose and the bandwidth both have equal values for all paths, therefore, they are ignored by the algorithm for now because there is no result from using them for comparison.

TABLE 6
DATA OF EACH PARAMETER FOR EACH PATH

| Solution/ Parameter | P. Loss | P. Delay | Cost | Bandwidth | Transmission Speed | Fitness $f(x)$ |
|---|---|---|---|---|---|---|
| X1 | 0 | 13 | 150 | 64 | 4 | 491.290000000000 |
| X2 | 0 | 29 | 300 | 64 | 7 | 794.250578512397 |
| X3 | 0 | 25 | 700 | 64 | 16 | 6672.84210526316 |

Looking at the other parameters, the first path has the lowest delay, then, the second path, then the third path. Next, considering the cost, the first path is the lowest again. Finally, for transmission speed, the first path is the best one again. As a result, the first path was chosen by DSO as the best solution.

The DSO will maintain it as the best solution, however, in case of any events that could impact the fitness of the solution, one of the reactions of the adaptive part of the algorithm will be implemented. The second-best solution to be chosen is the second path. Despite the fact that it has a higher delay than the third path but the cost difference is much higher that makes choosing the third path inefficient.

*6.3 AMBULANCE ROUTING*

If we consider traffic for an ambulance as an application for this algorithm, we can give the following example, if an ambulance is transferring a patient from one spot (in this scenario it is the University of Kurdistan-Hewler), which is located in the city center to an emergency hospital (in this scenario it's Rojawa Emergency hospital) located at the west side of the city, then, there are three possible roads to take (See Figures 13, 14, and 15);
1. Via Newroz road
2. Via Makhmour and Peshawa Qazi road
3. Via Mosul and Qazi Muhammed Road

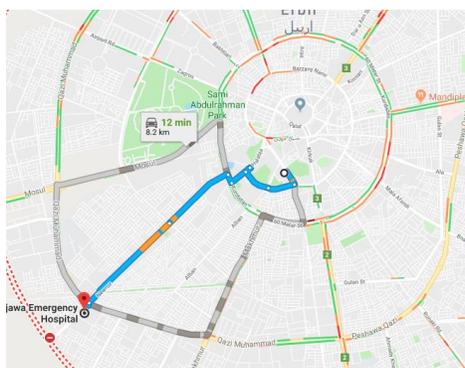
Figure 13: UKH to Rojawa Emergency Hospital via Newroz road, a road map (Google Maps, 2018)

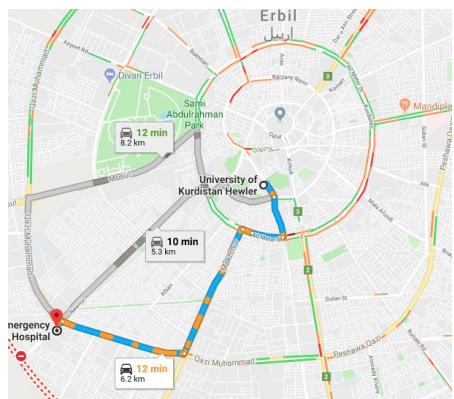
Figure 14: UKH to Rojawa Emergency Hospital via Makhmour and Peshawa Qazi road, a road map (Google Maps, 2018).

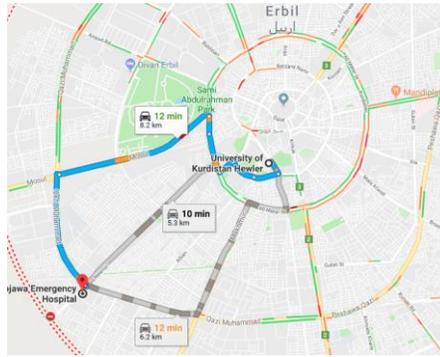
Figure 15: UKH to Rojawa Emergency Hospital via Mosul and Qazi Mohammed road, a road map (Google Maps, 2018).

There are three possible solutions that can solve the problem of which road to take, yet, there can only be one best solution that would give the best results. From a real-life evaluation of the three possible solutions based on the road condition, distance, cost, speed limits and how fast the ambulance can go on each road. The following can be stated; the speed limits are the same on all roads, the first choice has the best road condition and the ambulance can go faster on this road than the others, the distance of the first route is the lowest, then, the second rout, then, the third and the costs are moderate. The first route has the best road condition and the lowest number of road pumps, therefore, the ambulance can go the fastest on that route. Table 7 can sum up the parameters of each solution.

TABLE 7
The Assumed Parameters of each Possible Solution

| Solution/ Parameter | $X_{i1}$ | $X_{i2}$ | $X_{i3}$ | $X_{i4}$ | $X_{i5}$ | Fitness |
|---|---|---|---|---|---|---|
| | Road Condition | Distance | Cost | Speed Limit | Speed | |
| $X_{1j}$ | 1 | 1 | 2 | 0 | 1 | 0.275 |
| $X_{2j}$ | 2 | 3 | 3 | 0 | 4 | 0.172 |
| $X_{3j}$ | 2 | 4 | 4 | 0 | 3 | 0.169 |

From the fitness results shown in the table above, as we are going for the maximum fitness, we can see that the solutions can be sorted as $X_1$ is the best solution, then, $X_2$, then, $X_3$.
When these parameters are entered in the algorithm, the first solution had the best fitness so it is the best solution. To simulate this in the algorithm steps, this will be the first part of the algorithm and it will be as below:

Part I
1. The driver will check the map for possible solutions.
2. The Parameters of each solution will be collected.
3. The Fitness function will be executed on each solution to find the best one, in this scenario it is $X_1$.

Once the best solution is determined, then, it will be used, the reaction to the changes that can happen to the best solution can be shown as in the following, which will make the second part of the algorithm;

Part II
1. The traffic police patrols and their radio station can work as a congestion control mechanism, where they report any changes that can occur on the road. Such as accidents or traffic jams which might affect the fitness of the solution.
2. If any such events that impact the fitness of the best solution occur, as shown in Table 8, then one of the reactions will be implemented.

TABLE 8
CHANGES IN PARAMETERS OF POSSIBLE SOLUTION

| Solution/Parameter | $X_{i1}$ Road Condition | $X_{i2}$ Distance | $X_{i3}$ Cost | $X_{i4}$ Speed Limit | $X_{i5}$ Speed | Fitness |
|---|---|---|---|---|---|---|
| $X_{1j}$ | **3** | 1 | 2 | 0 | **3** | 0.275 |
| $X_{2j}$ | **1** | 3 | **2** | 0 | **2** | 0.172 |
| $X_{3j}$ | 2 | 4 | 4 | 0 | 3 | 0.169 |

The changes shown in red in Table 8 are, the road condition of the second possible solution has improved while one of the first solutions has decreased. The cost of the second possible solution has decreased too. The last change is the speed in the second possible solution has increased, on the contrary, it has decreased in the first solution. i.e. the ambulance can go faster on the second solution now than on the first solution. Therefore, the parameters of the possible solutions will be reevaluated and the best solution will become the second route as it has the highest fitness now. As a result, one of the reactions below will be taken to adopt this change in fitness.
- a. Run: stop using the first solution as the best solution and use the second solution as the best solution.
- b. Face and Suicide: the second solution will be used as the best solution until the first solution is back to its fitness. This is can be known through the continuous evaluation of the solutions using the fitness function.
- c. Face and Support: when a decrement in the fitness of the best solution is detected, we can avoid completely losing this solution by using the second-best solution as a supportive solution for the first best solution. i.e. the two solutions will be used for the same purpose in order to reduce the load on the first best solution which will give more chances to the ambulance to reach the hospital in the shortest period of time.
3. If there is no drop in the fitness of the best solution, then the normal operation is kept. That keeps using the current best solution and keep evaluating it against the other possible solutions in order to check its fitness.

The data of the roads are collected from google maps and the results we obtained from running the DSO is the same as the one in google maps. However, Google maps base their results on estimations and this makes the results unreliable sometimes as they don't reflect the real changes in the roads in terms of constructions, traffic, weather or any other events that may affect the fitness of the roads, see Figure 16. On the other hand, the DSO base it's results on the data collected by the traffic police. This makes the data more real-time. As a result, the solution given by the DSO is more real and reliable.

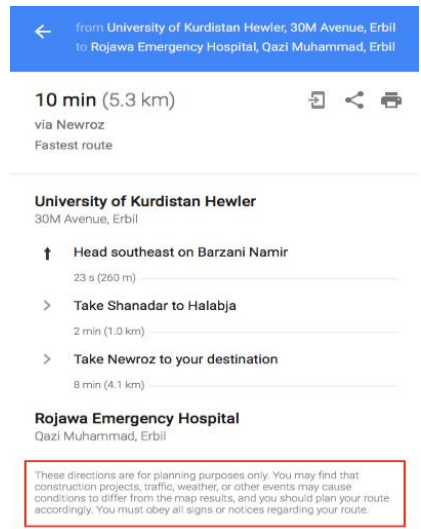

Figure 16: The warning message on the routes chosen (Google Maps, 2018).

## 7. CONCLUSION AND FUTURE WORK

In this research work, a novel Donkey and Smuggler Optimization algorithm was suggested. The DSO is inspired by very simple concepts and some of the searching behavior of donkeys, such as imitating the transporting behavior in terms of searching and selecting routes by donkeys in the real world. The novelty of this algorithm can be summarized in the following points:
- It is known that other algorithms have a randomized structure to find the best solution or creating more solutions. However, in the DSO, we have changed that. As it doesn't use the standard randomized structure and in addition to finding the best solution, it tries to maintain that solution.

- Unlike other algorithms that require parameter tuning and fixing to be adopted to solve real-world problems, this algorithm is easy to be implemented to a wide variety of problems due to its simple structure where it is parameter free and no derivations, as some algorithms require. I.e. it doesn't require any parameter tuning or fixing.
- As demonstrated in section 5: Performance evaluation, we can notice that the algorithm results overcome some of the optimization problems with very promising results. Also, in section 6.1: Travel salesman Problem, the algorithm could easily show superiority over the ACO in both, the verity of solutions found and their solidness as well as the execution time.

The DSO has two modes that are developed for implementing the searching and selecting routes. The modes are the Smuggler (non-adaptive) and Donkeys (adaptive). In the Smuggler mode, all the possible paths and the shortest path are discovered. Whereas in the Donkeys mode, the behaviors of the donkeys are exploited and explored to adapt to the changes in the best solution. The DSO is implemented to tackle three real-world problems: traveling salesman problem, packet routing, and ambulance routing. The experimental results concluded the following points:

1) It was established from the relevant literature that there isn't one global algorithm that can tackle capably almost all optimization problems. This will encourage researchers to make swarm intelligence extremely dynamic for scientific, engineering medical and business, military improvements and others.
2) The DSO algorithm is different from all other previous algorithms in optimization style.
3) The Smuggler mode determines the best solution among all workable solutions and the Donkey mode aims at maintaining the optimal solutions or returning to the optimal solution should the conditions be found.
4) Previous swarm intelligence algorithms use random selection processes for determining the best solution. I.e., if the global solution disappears for some reasons and for other reasons the global solution re-appears at later stages, then those algorithms are not devised to obtain the best solution quickly. On the contrary, the DSO does not depend on a random selection process. It instead uses a non-adaptive procedure to find and fix the optimum solution and then tries to maintain it through the adaptive part of the algorithm.
5) The parameters for any network would be the cost, bandwidth, delay, transmission speed, and packet loss. Yet, selecting any interior routing protocols that find the shortest path will not take all these parameters together into consideration. That is why a network was designed and configured using the OSPF (Open shortest path first) routing protocol and the pings were manually done through the possible paths to collect the required data.
6) The DSO will maintain the best solution; however, in case of any events that might impact the fitness of the best solution, then one of the reactions in the adaptive part of the algorithm, which are Run, Face and Suicide, and Face and Support will be implemented to optimize the system. Once the original best solution that has been set in the non-adaptive part is back to its ideal fitness, then the DSO will set it as the optimum solution again.
7) Compared to ACO in solving the TSP, the DSO could easily show superiority over the ACO in both, the verity of solutions found and their solidness as well as the execution time. The DSO, in one execution, gave four different possible solutions with an execution time being around 0.0385 seconds whereas, the ACO, with 16 executions and 100 iterations for each execution, gave unstable solutions with an execution time being around 0.11 to 0.13 seconds for each execution.
8) In the performance evaluation, the DSO was compared with five algorithms, VLE, PSO, GSA, DE, and WOA using ten benchmark testing functions. In the overall performance, DSO ranked 3$^{rd}$ amongst the six algorithms.
9) The experimental results of DSO on real-world problems were very promising. The results exhibited that the suggested DSO is appropriate to tackle other unfamiliar search spaces and complex problems like navigation and GPS.

As a future work, we would like to extend the performance evaluation section by including more benchmark functions as well as evaluating the algorithm with the CEC test functions. Furthermore, we will increase the number of dimensions and evaluate the performance of the algorithm with more decision variables and apply the algorithm to more real-world problems.

# Appendix 1:

The MATLAB source code for implementing the DSO algorithm on the TSP problem.

```matlab
tic;
clear all
clear min
Dis =[0 10 12 11 14; 10 0 13 15 8; 12 13 0 9 14 ; 11 15 9 0 16; 15 8 14 16 0];

figure()
names = {'path from city 1' ,'path from city 2' ,'path from city 3', 'path from city 4', 'path from city 5'};
namespossible = {'Possible path 1' ,'Possible path 2' ,'Possible path 3', 'Possible path 4'};

for i=1:5
    temp=Dis;
    temp(temp == 0 ) = NaN;
    temp(:,i) = NaN;
        out = min(temp(i,:)) ;
        [rowOfMin, colOfMin] = find(temp(i,:) == out) ;
        temp(:,colOfMin) = NaN;

        out1 = min(temp(colOfMin,:)) ;
        [rowOfMin1, colOfMin1] = find(temp(colOfMin,:) == out1) ;
        temp(:,colOfMin1) = NaN;

        out2=min(temp(colOfMin1,:));
        [rowOfMin2, colOfMin2] = find(temp(colOfMin1,:) == out2) ;
        temp(:,colOfMin2) = NaN;

        out3=min(temp(colOfMin2,:));
        [rowOfMin3, colOfMin3] = find(temp(colOfMin2,:) == out3) ;

        % showing the results on the console
        disp('city')
        disp(i)

        path=[i colOfMin colOfMin1 colOfMin2 colOfMin3 i]
        weight=out+out1+out2+out3+Dis(colOfMin3,i)

        % showing the results on the graph
        subplot(3, 2, i) ;
         plot(path) ;
         title([ names(i)  'weight' num2str(weight)]);
         ylabel('City','FontSize',12,'FontWeight','bold');
         grid on
         set(gca,'XTick',[])
         set(gcf,'units','points','position',[10,10,900,900])
end
execution_Time= toc
x = input( 'choose the departure city: ' );

 tic;
   %% finding the possible paths
   figure()
   for r=x:x

      n=1; % counter for the subplots

      for c=1:5
          temp=Dis;
          temp(temp == 0 ) = NaN;
          temp(:,r)=NaN;

         if ~isnan(temp(r,c))
             out = temp(r,c) ;
             [rowOfMin, colOfMin] = find(temp(r,:) == out) ;
             temp(:,colOfMin) = NaN;

             out1 = min(temp(colOfMin,:)) ;
             [rowOfMin1, colOfMin1] = find(temp(colOfMin,:) == out1) ;
             temp(:,colOfMin1) = NaN;

             out2=min(temp(colOfMin1,:));
```

```matlab
            [rowOfMin2, colOfMin2] = find(temp(colOfMin1,:) == out2) ;
            temp(:,colOfMin2) = NaN;

            out3=min(temp(colOfMin2,:));
            [rowOfMin3, colOfMin3] = find(temp(colOfMin2,:) == out3) ;
            temp(:,colOfMin3) = NaN;

            path=[x colOfMin colOfMin1 colOfMin2 colOfMin3 x]
            weight=out+out1+out2+out3+Dis(colOfMin3,x)

            temp(r,c)=NaN;

            subplot(2, 2, n) ;
            plot(path) ;
            title([ namespossible(n)  'weight' num2str(weight)]);
            ylabel('City','FontSize',12,'FontWeight','bold');
            grid on
            set(gca,'XTick',[])
            set(gcf,'units','points','position',[10,10,900,900])

            n=n+1;
        end

    end

    end
execution_Time2= toc;

  fprintf('exectution time of the first part = %f',execution_Time);
  fprintf('\n');
  fprintf('exectution time of the second part = %f',execution_Time2);
  fprintf('\n');
```